\documentclass[conference]{IEEEtran}
\IEEEoverridecommandlockouts

\makeatletter
\def\endthebibliography{%
	\def\@noitemerr{\@latex@warning{Empty `thebibliography' environment}}%
	\endlist
}
\makeatother
\usepackage{hyperref}
\hypersetup{draft}
\usepackage{verbatim}
\usepackage{textcomp}
\usepackage[pdftex]{graphicx}
\usepackage{multicol}
\usepackage{multirow}
\usepackage{caption}
\usepackage{subcaption}
\usepackage{array}
\usepackage{fixltx2e}
\begin{document}
\title{A Deep Transfer Learning-based Edge Computing Method for Home Health Monitoring}

\author{\IEEEauthorblockN{Abu Sufian}
\IEEEauthorblockA{\small{\textit{Department of Computer Science}}\\
\small{\textit{University of Gour Banga}}\\
\small{Malda, India}\\
\small{Email: sufian@ieee.org}}
\and
\IEEEauthorblockN{Changsheng You}
\IEEEauthorblockA{\small{\textit{Department of Electrical and Computer  Engineering}}\\
\small{\textit{National University of Singapore}}\\
 \small{Singapore 117583, Singapore}\\
\small{eleyouc@nus.edu.sg}}
\and
\IEEEauthorblockN{Mianxiong Dong}
\IEEEauthorblockA{\small{\textit{Department of Sciences and Informatics}}\\
	\small{\textit{Muroran Institute of Technology}}\\
	\small{Hokkaido, Japan}\\
	\small{mx.dong@csse.muroran-it.ac.jp}}}

\IEEEoverridecommandlockouts
\IEEEpubid{\makebox[\columnwidth]{
		xxx-x-xxxx-xxxx-x/21/\$xx.xx~\copyright~2021 XXXX \hfill}
	 \hspace{\columnsep}\makebox[\columnwidth]{ }}
	
\maketitle
\renewcommand\IEEEkeywordsname{Keywords}

\begin{abstract}
    The health-care gets huge stress in a pandemic  or epidemic situation. Some diseases such as COVID-19 that causes a pandemic is highly spreadable from an infected person to others. Therefore, providing health services at home for non-critical infected patients with isolation shall assist to mitigate this kind of stress. In addition, this practice is also very useful for monitoring the health-related activities of elders who live at home. The home health monitoring,  a continuous monitoring of a patient or elder at home using visual sensors is one such non-intrusive sub-area of health services at home. In this article, we propose a transfer learning-based edge computing method for home health monitoring. Specifically, a pre-trained convolutional neural network-based model can leverage edge devices with a small amount of ground-labeled data and fine-tuning method to train the model.  Therefore, on-site computing of visual data captured by RGB, depth, or thermal sensor could be possible in an affordable way. As a result, raw data captured by these types of sensors is not required to be sent outside from home. Therefore, privacy, security, and bandwidth scarcity shall not be issues. Moreover, real-time computing for the above-mentioned purposes shall be possible in an economical way.      
\end{abstract}
    
\section{Introduction}
In India, only 1.9 millions hospital beds in all kind hospitals are currently available for population around 1.35 billion \cite{Kapoor20}; that is, only 1.4 beds per 1000 peoples. This situation is also not far better in other countries \cite{blavin2020hospital}. In addition, those countries that are comparably on top of that list also may not be able to cope with the challenges arising from a pandemic. Therefore, home health services need to be improved to cope with a pandemic or epidemic such as COVID-19. Moreover, as the parentage of aged people(elders) is increasing steadily \cite{ageing}, so home health services, are also very useful health practice for elders who live at home. \par  
As Artificial Intelligence (AI) is augmenting human capabilities for many human-centred tasks \cite{shneiderman2020human, he2019practical, xu2019sceh}. Therefore, AI could also assist home health services in many ways \cite{fritz2019nurse, yang2020homecare}.  Automated patient or elders monitoring (in short we are calling it `Home Health Monitoring') one such non-intrusive and economical sub-area of these services; these sub-area may include activity monitoring, sleep monitoring, respiration monitoring, fall detection, facial expression understanding, speech recognition, hand hygienic practice monitoring,  etc. \cite{malasinghe2019remote}. For these kind of tasks, deep learning (DL) and computer vision (CV) are very effective as studied in \cite{sathyanarayana2018vision, luo2018computer, Yang_2020}. But DL especially for tasks of CV, required GPU-enabled computing devices \cite{feng2019computer}, which may not be available for every household. To address this issue, one approach is to leverage cloud computing\cite{yang2017big} technique, where data needs to be sent to a remote cloud server for processing outside from home. But in this case, privacy, security and bandwidth scarcity are big issues and real-time computing may not be possible \cite{sufian21CVIoT}. These disincentives motivate to use the new technology of Edge Computing (EC) \cite{satyanarayanan2017emergence}. EC could be used to compute data of home health monitoring inside the home or house. However, some challenges also exists as edge devices(ED) are generally small and have low computing capabilities \cite{shi2016edge}. In addition, the DL-based model usually takes a large amount of data which is also a big challenge for health sectors \cite{shen2017deep}. \par
In this article, we propose a deep transfer learning-based edge computing method for home health monitoring (TL-EC-HM). Here, we consider a transfer learning approach, where a pre-trained Convolutional Neural Network (CNN \cite{ghosh2020fundamental})-based model which is trained with its available dataset, may use in ED with fine-tuning using a small amount of ground labeled dataset.
\begin{figure}[!htb]
	\centering
	\includegraphics[width=.95\linewidth]{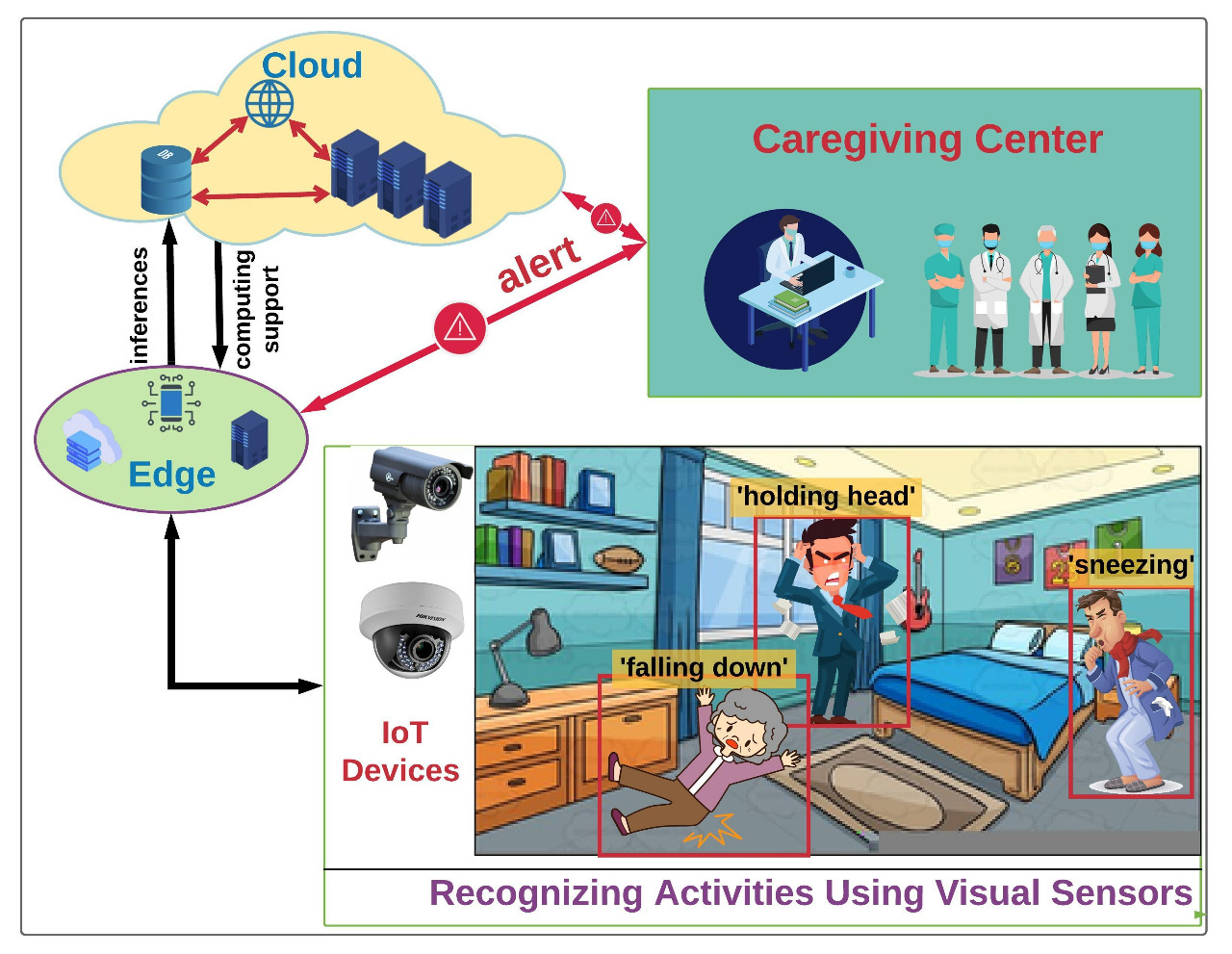}
	\caption{A working scenario of TL-EC-HM}
	\label{fAM}
\end{figure}      
In this way, it would take much less computing resources and the required on-site visual computing shall be possible at an ED. Therefore, mitigation of the above-mentioned challenges shall be possible.     
A possible working scenario of TL-EC-HM is depicted in Fig. \ref{fAM}, where how a caregiver center, cloud server, ED, and IoT device(sensor) are connected to each other to form a system is shown.
  The highlights of this article are listed as below:
\begin{itemize}
	\item We provide a study on health and activity monitoring for patient as well as elders at their home for mitigating health crisis.
	\item We propose a method (TL-EC-HM) based on  DTL and EC for home health monitoring.
	\item We analyze the proposed privacy-preserving TL-EC-HM for on-site visual computing.
	\item We provide some future research directions.
\end{itemize}
The rest of this article is organized as follows: In Sec. \ref{relatedWorks}, recent related works are mentioned. Discussion of the proposed method is in Sec. \ref{description}. An analysis of the TL-EC-HM is in Sec. \ref{analysis} whereas future scopes of the method are given in Sec. \ref{FS}. Finally, conclusions of the article are provided in Sec. \ref{conclusion}. 

\section{Recent Related Works}
\label{relatedWorks}
As we are proposing a new method for home health monitoring \cite{wang2020elderly}, so existing related studies are few. However, many sensor based activity recognition approaches \cite{chen2019sensor} may exploit for these purposes. Some of those recent works that are related to the objective of our work are reported below: \par 
In a study \cite{luo2018computer}, Z. Luo et al. applied DL and CV to analyze activities of seniors for health monitoring. In that non-instructive approach, they suggest automated analytical inference of activities of daily living, such as sleeping, walking, sitting, standing, etc. In their pilot study, they used depth and thermal sensors for preserving privacy of seniors. In another works, E. Chou et al.  proposed an action recognition study with low resolution depth images \cite{chou2018privacy}. Here, they downsampled depth images to preserve privacy for two healthcare surveillance scenarios, those are hand-hygiene compliance and ICU activity. They applied a privately trained skip connected CNN  model to enhance the inference from image frames. In a related study \cite{deng2018design}, F. Deng et al. proposed a non-contact sleep monitoring system using infrared and motion sensor. Their  combined infrared video frames captured by five infrared cameras installed with different orientation.   After that they applied a machine learning techniques to get inference to classifying respiration, head posture and body posture. In a study \cite{xue2018vision}, D. Xue et al. proposed vision based senior care using gait analysis at home for similar purposes. \par In a study \cite{pfarr2019avatar}, J. Pfarr et al. proposed an avatar based patient monitoring. In their peripheral vision-based monitoring system assist respective caregiver centre by sending information if any anomalies is detected. In another study \cite{ahmedt2019understanding}, D. Ahmedt-Aristizabal et al. proposed a vision-based analysis of seizure disorders through understanding patients’ behaviour. The authors proposed two marker free DL models, one is landmark-based and the other one is region-based for their work comparison.  On the other hand, E. Dolatabadi proposed a feasibility study \cite{dolatabadi2019feasibility} on a vision-based sensor for longitudinal monitoring of mobility in the elders with dementia. They used vision based sensor pose tracking sensors for analysis of gait over time for spatio temporal measures dementia of the elders. L. A. Zavala-Mondragon et al. proposed CNN-SkelPose \cite{zavala2019cnn}, which is a CNN-based skeleton model estimation method for clinical applications. Here, the authors used depth images for skeleton  estimation for patient monitoring. In a study \cite{yeung2019computer}, S. Yeung et al. proposed a CV and DL-based detection of patient mobilization activities in an ICU. They collected privacy-safe a depth video dataset from a hospital containing the activities such as moving  patients into and out of a bed or chair. Here, the authors used state-of-the-art DL model to quantify the activities.  Y. J. Park et al. proposed Deep-cARe \cite{park2019deep}, a projection-based augmented reality (PAR)  with DL for elders care at home. They used bidirectional PAR and DL to get contextual awareness of pose estimation, face recognition and object detection for elders at home.  
In another study \cite{kim2019vision}, K. Kim et.al proposed a vision based human activity recognition system using depth silhouettes for monitoring the elderly resident in a smart home. In their depth sensor-video-based system,  it utilizes skeleton joints features from data, and that recognize activities of the elders at home. They used depth map to track human silhouettes and body joints information before feeding it into a hidden Markov model to recognize human activities. \par
M. Buzzelli et al. proposed a CV-based monitoring dataset and model for the elder at home \cite{buzzelli2020vision}. Here, they created a dataset (called it ALMOND) and proposed a DL-based model with three baseline accuracies for three different task, namely basic poses, on alerting situations, and daily life actions. In another study \cite{khraief2020elderly}, C. Khraief et al. proposed a multi-stream deep convolutional networks for fall detection. They created a CNN architecture with four separate CNN streams, one for each modality of RGB-D camera images. In their model, first modality is used for illumination variations, second one for human shape variations, and last two for more discriminate information of motion. Weighted score of each modality are fusion to get system performance. \par   
The above mentioned works have been contributed largely but they focus only CV, DL, and dataset. None of the works focused these techniques with EC for on-site visual computing. Specifically, no work found that used DTL in ED for home health monitoring.
     
\section{Methods of TL-EC-HM}
\label{description}
The proposed TL-EC-HM is works with association of some technical components such as sensor, EC, and DTL. These are directly relevant to this work, therefore these are first briefly mentioned, then the propose pipeline is discussed.    

\subsection{Type of Sensor May Used}
\label{sensors}
In order to do home health monitoring, mainly five type of sensors, i.e., RGB, Depth, Thermal(Inferred), Sound, and Wearable sensors may be used. As our proposed TL-EC-HM focuses on non-intrusive vision-based monitoring, so we pay attention to compute spatial data captured by first three type sensors that are visual in nature. Among these three types of sensors, RGB could give more details. On the other hand, depth and thermal sensors may be used for hiding privacy with comparably less details. As we are proposing an EC strategy, privacy is not a big issue compared to fully cloud computing-based strategies.
In addition, we can use multimodal images captured by multiple sensors.
A multimodal combination could also be made by depth, thermal, and other ambient sensors in order to acquire the strengths of each sensor.  

\subsection{Edge Computing(EC)}
Computing is the necessary functionalities to make desired inference from the data sensed by sensors. As mentioned, EDs that embedded with sensors have very limited computing capabilities, so, cloud or sometimes fog computing may be used. But in both cases, data needs to be sent outside of the home, so, privacy, security, latency, etc. become a huge problem and that leads towards the edge computing  \cite{satyanarayanan2017emergence}. \par
EC, a computing technique where computation performs at or near the devices, has made them useful. EC shall be more powerful when benefit of DL could adopted in edges \cite{ li2018learning, chen2019deep, wang2020convergence}. But here computational load and dependency of large amount of labeled data to train a DL model are more challenging. Therefore, transfer learning or few shot learning \cite{wang2020generalizing} are very useful in EC.   

\subsection{Deep Transfer Learning(DTL)}
\label{DTL}
Modern AI technologies largely depends on DL \cite{lecun2015deep}.
To train a DL model from scratch, it requires a massive amount of training data whereas medical data is not easily available \cite{esteva2019guide}. Moreover, the DL-based model usually requires large computing power such as GPU-enabled high resource consumption machine which is a big challenge for EC \cite{chen2019deep}. On the other hand, DTL is a technique which uses features of an already trained DL model to solve new task with required fine-tuning \cite{tan2018survey}. DTL significantly reduces the requirement for training data and computing resources for a target domain-specific task. \par
Therefore, DTL, that is DL-based Transfer Learning technique is used to overcome these challenges.  Here, a DL-based model is trained on another large available dataset of related domain at cloud server or GPU-enabled machine for feature extraction. After that, the pre-trained model is used at edges with fine-tuning with some actual ground labeled data of monitoring task. Therefore, DTL makes ED more intelligent in a affordable way to assist for mitigating health crisis \cite{sufian2020survey} through home health monitoring.\par
\begin{figure*}[h]
	\centering        
	\includegraphics[width=.9\textwidth]{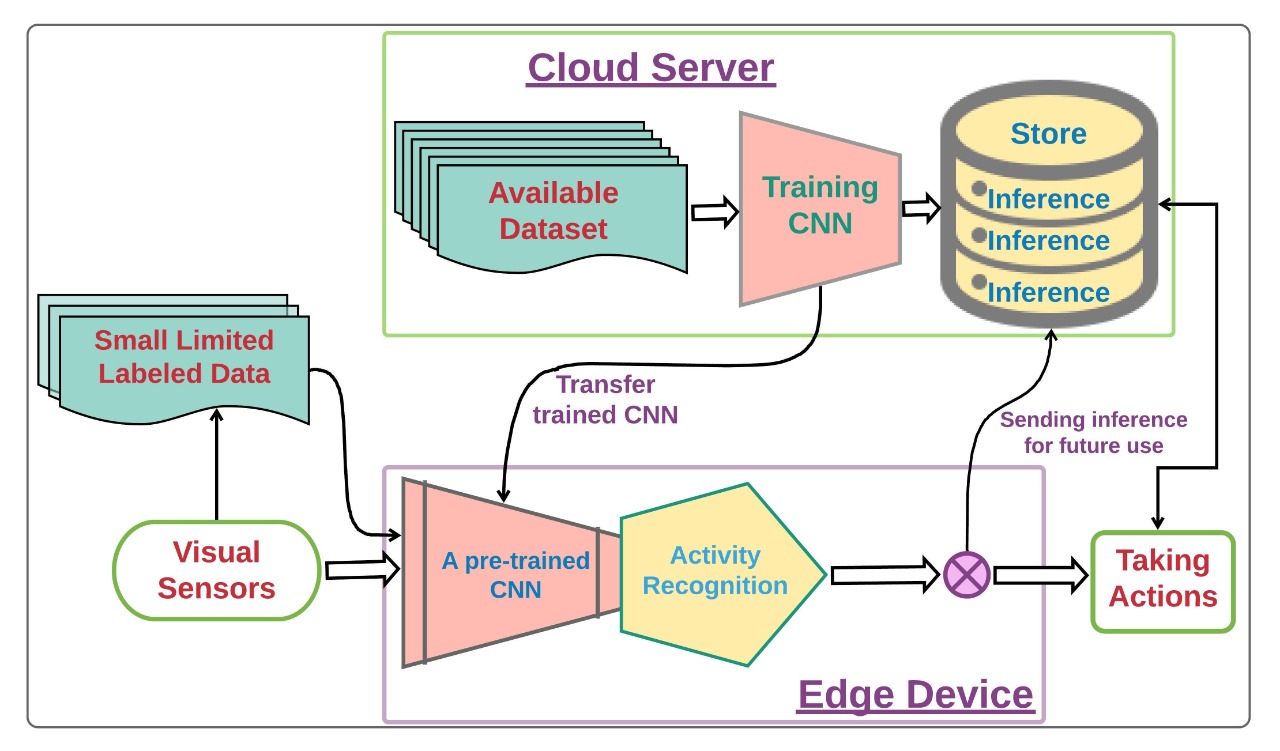}
	\caption{A graphical look of proposed pipeline of TL-EC-HM}
	\label{pipline}
\end{figure*}
\subsection{Pipeline of TL-EC-HM}
As mentioned, ED has very limited computing capabilities. Thus, our proposed TL-EC-HM gets assistance of a cloud server. Here, a suitable CNN-based activity recognition model is trained in a GPU-enabled cloud server with available dataset. Then the pre-trained model is sent to EDs that are installed in home. An ED captured image or video streams use one or more visual sensors as mentioned in Sec. \ref{sensors}. In the EDs, a small amount of ground labeled-data is created for fine-tuning the pre-trained model. After completion of fine-tuning, the ED is ready to perform EC on continuous sense data capture by installed sensor(s) for home health monitoring. Here, EDs recognize activities with help that pre-trained CNN classifier, and this inference is forwarded to caregiver centre as well as to cloud server. The associate caregiver center takes appropriate actions based on current inference forwarded by ED and information that are stored in a cloud storage. The proposed pipeline is graphically shown in Fig. \ref{pipline}. There are four major parts of the methods those are discussed below. 

\subsubsection{Training CNN}         
As mentioned, an activity recognition CNN needs to be trained using a suitable targeted task related dataset in a cloud server or high computing machine.  A cloud server that has capabilities to run an activity recognizing CNN with  GPU, energy and other required resources has been considered in this method. \par
Let's suppose  a C-multi-way classification
task with a available training dataset: $$dataTrain  = \{(x_i, y_i): i= 1 \mbox{ to }  N\}$$
where $x_i$ is a i-th video frame 
of an activity recognizing video data, $y_i$ is an integer in [1, C] represent activities as the class label, and $N$ is the number of frames. The video could be RGB or thermal or depth, or even a multimodal combination depth and thermal for extra privacy. Here, the objective is to train the required CNN with a hypothesis $h$ in (\ref{hyp}), so that it can predict the exact pair $y_i$ of frame $x_i$ from test set:$$ 
dataTest  = \{(x_i, y_i ): i= 1 \mbox{ to }  N^\prime \}$$
where $N^\prime$ is the number of test frames in test video.
\begin{equation}
h = \arg \min \frac{1}{N^\prime} \sum_{(x_i, y_i) \in dataTest}Error(Loss(\widehat{y_i}, y_i))
\label{hyp}
\end{equation}

Here, $\widehat{y_i}$ denotes the predicted value of input $x_i$ and $loss$ is the softmax cross entropy loss as (\ref{loss}).
\begin{equation}
Loss(\widehat{y_i}, y_i) = -  \sum_{j=1}^{C} y_{i=j}\log \widehat{y_{j,i}}
\label{loss}
\end{equation}
where  $\widehat{y_{j,i}}$ denotes the output probability for class $j$ on input $x_i$ and it is obtained by the softmax function in (\ref{soft})
\begin{equation}
\widehat{y_{j,i}} = \frac{e^{\widehat{y_i}}}{\sum_{j=1}^{C} e^{\widehat{y_{j,i}}}}
\label{soft}
\end{equation}
After that, the trained CNN is pushed to the ED which is installed in a home or house.  

\subsubsection{Fine-tuning of a Pre-trained CNN}
This is the first computing step at ED. First, a small amount ($n$ number for training and $n^\prime$ for testing) of real data captured by deployed sensors with ground truth label is created for train and test as:
$$dataTrain^\prime  = \{(x_i^\prime, y_i^\prime): i= 1 \mbox{ to }  n\}$$
$$dataTest^\prime   = \{(x_i^\prime, y_i^\prime): i= 1 \mbox{ to }  n^\prime \}$$
Then the trained CNN is fine-tuned using this dataset to make the ED ready to perform required EC. Here also set of activities are selected from 1 to $C^\prime$. The same set of operations as mentioned in (\ref{hyp}), (\ref{loss}) and (\ref{soft}) are performed for fine-tuning of last few layers (depending on task and ED) of the trained CNN-based model. After this steps, EDs shall be ready to recognized action on frame stream of running video.      

\subsubsection{Running Activity Recognizing}
After preparing the CNN-based activity recognizing model, the overlapping frames are extracted window-wise  from running video stream. Each window passes through the CNN to get window level probability score. Then it averages to get frame level score. Suppose a window size is [$t1$ to $t2$] and if $K$ overlapping frames are taken, then frame level probability score is calculated by (\ref{fscore}). 
\begin{equation}
	\widehat{y_i}^\prime = \frac{1}{K}\sum_{\forall j \in K} \widehat{y_j}
	\label{fscore}
\end{equation}
where $\widehat{y_i}^\prime$ and $\widehat{y_i}$ are  probability scores of mean and each frame of the window [t1, t2]. \par
This frame-level probability that is softmax score is used to  recognized frame level running activities. These are used to calculate the frame-level mean average precision for judgment and drawn inference. Then from ED it forwarded to caregivers center and cloud server for taking appropriate action and storing respectively.  

\subsubsection{Taking Actions}
Caregiver center receives categorical inference from EDs. The categories may be serious alert, required some service, etc.  Based on these categories, caregiver center takes appropriate action. The actions could be service related, could be consultant with doctors, etc. The requirement caregiver may get stored information from cloud storage for long term assessment of the patient or the elder. \par 
Important point is that, with this method, only inferences (not raw data) go outside from home, this adds extra protection of privacy which will motivate the patient or elder to use home health monitoring services. 
\begin{figure}[!hbt]	
	\centering
	\includegraphics[width=0.95\linewidth]{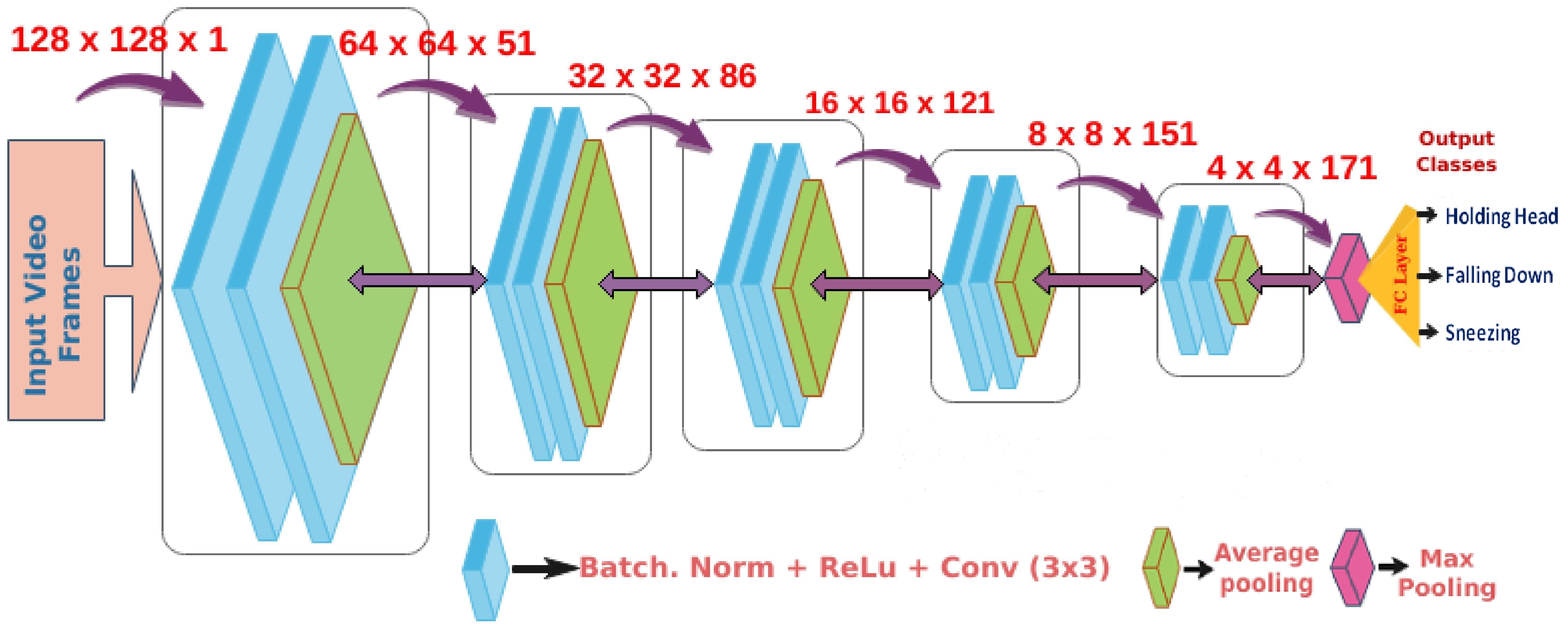}
	\caption{Architecture and number of parameters of our CNN.}
	\label{F_expcnn}       
\end{figure}
\section{An Analysis of Proposed TL-EC-HM}
\label{analysis} 
\begin{figure}[!hbt]	
	\centering
	\includegraphics[width=0.95\linewidth]{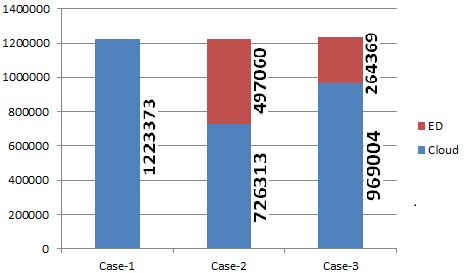}
	\caption{The number of parameters sharing for training.}
	\label{F_Tlcase}       
\end{figure}

The objective of this analysis to show the feasibility of the propose method. This gives an estimate on how only a few parameters are only required to trained for fine-tuning for the purpose if DTL technique used. In this TL-EC-HM, we have taken the CNN shown in Fig. \ref{F_expcnn} to reuse a pre-trained CNN with required fine-tuning. We can train the above mentioned CNN model with a suitable dataset at a cloud server using high computing resources, and then push the trained model into an ED which is installed at a home. In the ED, fine-tune the last few layers of the CNN with a ground-labeled training data using  a DTL technique. This training process in ED includes the aligning of output classes to required classes (in Fig. \ref{F_expcnn} it three), and making some front layers to be frozen except the last few layers. The number of freezing layers will depend on the trade-off between the accuracy and computing power of EDs. Fig.   \ref{F_Tlcase} shows the total number of parameters training in cloud server as well as in ED in three different observing cases:-
\begin{itemize}
	\item \textbf{Case 1:} The total number of parameters of the CNN in Fig. \ref{F_expcnn} is 1223373 that are trained in cloud server(typical cloud computing). 
	\item \textbf{Case 2:} Block-1 to block-4 of the CNN \ref{F_expcnn} are frozen (Our first DTL-based EC observation).
	\item \textbf{Case 3:} Block-1 to block-4 as well as the first $BatchNorm\rightarrow ReLU\rightarrow ConV2d(3\times3) $ layers of block-5 are frozen (Our second DTL-based EC observation).
\end{itemize}
Through our proposed method, training of parameters could be done by Case-2 or Case-3, for that parameters sharing is shown graphically in Fig. \ref{F_Tlcase}. Here, Case-1 shows that entire parameters need to be trained at the cloud server as cloud computing if DTL is not used. On the other hand, one could attempt to train all parameters in the ED which is not feasible for a low resources-based ED as discussed. For Case-2, 497000 out of 1223373 parameters are needed to fine tuned. For Case-3, only 264369 parameters out of 1223373 are needed to fine tuned, that is here only around 21\%. Therefore, the Case-3 is quite feasible for a tiny ED to run EC.   \par
This discussion gives an estimated idea that how the proposed TL-EC-HM needs less number of parameters to train(fine-tuning) a CNN-based model. That is, an entire CNN-based model may not be possible to run in EDs whereas TL-EC-HM based methods can work. 

\section{Future Scopes}  
\label{FS}
The proposed method has large scopes of applicability with further investigation. A pilot study by this method including dataset creation, system setup, and data analytics is felt immediate next phase of the study.  \par
 The main required components of this method are: A CNN-based action recognition model, a suitable available dataset for training the model, a small set of ground labeled dataset for fine-tuning, and instruments including IoT based visual sensors-enabled ED, cloud server, caregiver centre, home space,  etc. Using this setup, a pilot study shall be carried out. Before pilot study, one may choose a simulation study.  Moreover, this visual sensor based monitoring could also merge with other ambient sensors to add more features in this home health monitoring.    
 
\section{Conclusion}
\label{conclusion}
To mitigate the health crises in a pandemic or to take care elders in an affordable way, home health monitoring would be very beneficial. In this article, we have proposed a computer vision-based method where a deep transfer learning is used in edge devices as edge computing. In this approach, the raw visual data continuously capture by visual sensor(s) is not required to be sent outside of home. Therefore, privacy, data security as well as latency are not big issues. 

\section*{Acknowledgment}
This work is partially supported by JSPS KAKENHI Grant Numbers JP20F20080.

\bibliographystyle{IEEEtran}

\bibliography{OurBib}

\begin{thebibliography}{10}
\providecommand{\url}[1]{#1}
\csname url@samestyle\endcsname
\providecommand{\newblock}{\relax}
\providecommand{\bibinfo}[2]{#2}
\providecommand{\BIBentrySTDinterwordspacing}{\spaceskip=0pt\relax}
\providecommand{\BIBentryALTinterwordstretchfactor}{4}
\providecommand{\BIBentryALTinterwordspacing}{\spaceskip=\fontdimen2\font plus
\BIBentryALTinterwordstretchfactor\fontdimen3\font minus
  \fontdimen4\font\relax}
\providecommand{\BIBforeignlanguage}[2]{{%
\expandafter\ifx\csname l@#1\endcsname\relax
\typeout{** WARNING: IEEEtran.bst: No hyphenation pattern has been}%
\typeout{** loaded for the language `#1'. Using the pattern for}%
\typeout{** the default language instead.}%
\else
\language=\csname l@#1\endcsname
\fi
#2}}
\providecommand{\BIBdecl}{\relax}
\BIBdecl

\bibitem{Kapoor20}
G.~Kapoor, A.~Sriram, J.~Joshi, A.~Nandi, and R.~Laxminarayan, ``Covid-19 in
  india : State-wise estimates of current hospital beds, intensive care unit
  (icu) beds and ventilators,'' in \emph{CDDEP, Princeton University}, 2020.

\bibitem{blavin2020hospital}
\BIBentryALTinterwordspacing
F.~Blavin and D.~Arnos, ``Hospital readiness for covid-19,'' 2020. [Online].
  Available:
  \url{https://www.urban.org/sites/default/files/publication/101864/hospital-readiness-for-covid-19_2.pdf}
\BIBentrySTDinterwordspacing

\bibitem{ageing}
\BIBentryALTinterwordspacing
U.~Nations, ``World population ageing report,'' 2019. [Online]. Available:
  \url{https://www.un.org/en/sections/issues-depth/ageing/}
\BIBentrySTDinterwordspacing

\bibitem{shneiderman2020human}
B.~Shneiderman, ``Human-centered artificial intelligence: Reliable, safe \&
  trustworthy,'' \emph{International Journal of Human--Computer Interaction},
  vol.~36, no.~6, pp. 495--504, 2020.

\bibitem{he2019practical}
J.~He, S.~L. Baxter, J.~Xu, J.~Xu, X.~Zhou, and K.~Zhang, ``The practical
  implementation of artificial intelligence technologies in medicine,''
  \emph{Nature medicine}, vol.~25, no.~1, pp. 30--36, 2019.

\bibitem{xu2019sceh}
C.~Xu, M.~Dong, K.~Ota, J.~Li, W.~Yang, and J.~Wu, ``Sceh: smart customized
  e-health framework for countryside using edge ai and body sensor networks,''
  in \emph{2019 IEEE global communications conference (GLOBECOM)}.\hskip 1em
  plus 0.5em minus 0.4em\relax IEEE, 2019, pp. 1--6.

\bibitem{fritz2019nurse}
R.~L. Fritz and G.~Dermody, ``A nurse-driven method for developing artificial
  intelligence in “smart” homes for aging-in-place,'' \emph{Nursing
  outlook}, vol.~67, no.~2, pp. 140--153, 2019.

\bibitem{yang2020homecare}
G.~Yang, Z.~Pang, M.~J. Deen, M.~Dong, Y.~Zhang, N.~H. Lovell, and A.~M.
  Rahmani, ``Homecare robotic systems for healthcare 4.0: Visions and enabling
  technologies,'' \emph{IEEE Journal of Biomedical and Health Informatics},
  2020.

\bibitem{malasinghe2019remote}
L.~P. Malasinghe, N.~Ramzan, and K.~Dahal, ``Remote patient monitoring: a
  comprehensive study,'' \emph{Journal of Ambient Intelligence and Humanized
  Computing}, vol.~10, no.~1, pp. 57--76, 2019.

\bibitem{sathyanarayana2018vision}
S.~Sathyanarayana, R.~K. Satzoda, S.~Sathyanarayana, and S.~Thambipillai,
  ``Vision-based patient monitoring: a comprehensive review of algorithms and
  technologies,'' \emph{Journal of Ambient Intelligence and Humanized
  Computing}, vol.~9, no.~2, pp. 225--251, 2018.

\bibitem{luo2018computer}
Z.~Luo, J.-T. Hsieh, N.~Balachandar, S.~Yeung, G.~Pusiol, J.~Luxenberg, G.~Li,
  L.-J. Li, N.~L. Downing, A.~Milstein \emph{et~al.}, ``Computer vision-based
  descriptive analytics of seniors’ daily activities for long-term health
  monitoring,'' \emph{Machine Learning for Healthcare (MLHC)}, vol.~2, 2018.

\bibitem{Yang_2020}
X.~Yang, X.~Ren, M.~Chen, L.~Wang, and Y.~Ding, ``Human posture recognition in
  intelligent healthcare,'' \emph{Journal of Physics: Conference Series}, vol.
  1437, p. 012014, jan 2020.

\bibitem{feng2019computer}
X.~Feng, Y.~Jiang, X.~Yang, M.~Du, and X.~Li, ``Computer vision algorithms and
  hardware implementations: A survey,'' \emph{Integration}, 2019.

\bibitem{yang2017big}
C.~Yang, Q.~Huang, Z.~Li, K.~Liu, and F.~Hu, ``Big data and cloud computing:
  innovation opportunities and challenges,'' \emph{International Journal of
  Digital Earth}, vol.~10, no.~1, pp. 13--53, 2017.

\bibitem{sufian21CVIoT}
A.~Sufian, E.~Alam, A.~Ghosh, F.~Sultana, , D.~De, and M.~Dong, ``Deep learning
  in computer vision through mobile edge computing for iot,'' in \emph{Mobile
  Edge Computing}.\hskip 1em plus 0.5em minus 0.4em\relax Springer, 2021, p. in
  press.

\bibitem{satyanarayanan2017emergence}
M.~Satyanarayanan, ``The emergence of edge computing,'' \emph{Computer},
  vol.~50, no.~1, pp. 30--39, 2017.

\bibitem{shi2016edge}
W.~Shi, J.~Cao, Q.~Zhang, Y.~Li, and L.~Xu, ``Edge computing: Vision and
  challenges,'' \emph{IEEE internet of things journal}, vol.~3, no.~5, pp.
  637--646, 2016.

\bibitem{shen2017deep}
D.~Shen, G.~Wu, and H.-I. Suk, ``Deep learning in medical image analysis,''
  \emph{Annual review of biomedical engineering}, vol.~19, pp. 221--248, 2017.

\bibitem{ghosh2020fundamental}
A.~Ghosh, A.~Sufian, F.~Sultana, A.~Chakrabarti, and D.~De, ``Fundamental
  concepts of convolutional neural network,'' in \emph{Recent Trends and
  Advances in Artificial Intelligence and Internet of Things}.\hskip 1em plus
  0.5em minus 0.4em\relax Springer, 2020, pp. 519--567.

\bibitem{wang2020elderly}
X.~Wang, J.~Ellul, and G.~Azzopardi, ``Elderly fall detection systems: A
  literature survey,'' \emph{Front. Robot. AI}, vol.~7, p.~71, 2020.

\bibitem{chen2019sensor}
L.~Chen and C.~D. Nugent, ``Sensor-based activity recognition review,'' in
  \emph{Human Activity Recognition and Behaviour Analysis}.\hskip 1em plus
  0.5em minus 0.4em\relax Springer, 2019, pp. 23--47.

\bibitem{chou2018privacy}
E.~Chou, M.~Tan, C.~Zou, M.~Guo, A.~Haque, A.~Milstein, and L.~Fei-Fei,
  ``Privacy-preserving action recognition for smart hospitals using
  low-resolution depth images,'' \emph{Machine Learning for Health (ML4H)
  Workshop at NeurIPS 2018, Montréal, Canada.}, 2018.

\bibitem{deng2018design}
F.~Deng, J.~Dong, X.~Wang, Y.~Fang, Y.~Liu, Z.~Yu, J.~Liu, and F.~Chen,
  ``Design and implementation of a noncontact sleep monitoring system using
  infrared cameras and motion sensor,'' \emph{IEEE Transactions on
  Instrumentation and Measurement}, vol.~67, no.~7, pp. 1555--1563, 2018.

\bibitem{xue2018vision}
D.~Xue, A.~Sayana, E.~Darke, K.~Shen, J.-T. Hsieh, Z.~Luo, L.-J. Li, N.~L.
  Downing, A.~Milstein, and L.~Fei-Fei, ``Vision-based gait analysis for senior
  care,'' \emph{arXiv preprint arXiv:1812.00169}, 2018.

\bibitem{pfarr2019avatar}
J.~Pfarr, M.~T. Ganter, D.~R. Spahn, C.~B. Noethiger, and D.~W. Tscholl,
  ``Avatar-based patient monitoring with peripheral vision: a multicenter
  comparative eye-tracking study,'' \emph{Journal of medical Internet
  research}, vol.~21, no.~7, p. e13041, 2019.

\bibitem{ahmedt2019understanding}
D.~Ahmedt-Aristizabal, S.~Denman, K.~Nguyen, S.~Sridharan, S.~Dionisio, and
  C.~Fookes, ``Understanding patients’ behavior: Vision-based analysis of
  seizure disorders,'' \emph{IEEE journal of biomedical and health
  informatics}, vol.~23, no.~6, pp. 2583--2591, 2019.

\bibitem{dolatabadi2019feasibility}
E.~Dolatabadi, Y.~X. Zhi, A.~J. Flint, A.~Mansfield, A.~Iaboni, and B.~Taati,
  ``The feasibility of a vision-based sensor for longitudinal monitoring of
  mobility in older adults with dementia,'' \emph{Archives of gerontology and
  geriatrics}, vol.~82, pp. 200--206, 2019.

\bibitem{zavala2019cnn}
L.~A. Zavala-Mondragon, B.~Lamichhane, L.~Zhang, and G.~de~Haan,
  ``Cnn-skelpose: a cnn-based skeleton estimation algorithm for clinical
  applications,'' \emph{Journal of Ambient Intelligence and Humanized
  Computing}, pp. 1--12, 2019.

\bibitem{yeung2019computer}
S.~Yeung, F.~Rinaldo, J.~Jopling, B.~Liu, R.~Mehra, N.~L. Downing, M.~Guo,
  G.~M. Bianconi, A.~Alahi, J.~Lee \emph{et~al.}, ``A computer vision system
  for deep learning-based detection of patient mobilization activities in the
  icu,'' \emph{NPJ digital medicine}, vol.~2, no.~1, pp. 1--5, 2019.

\bibitem{park2019deep}
Y.~J. Park, H.~Ro, N.~K. Lee, and T.-D. Han, ``Deep-care: Projection-based home
  care augmented reality system with deep learning for elderly,'' \emph{Applied
  Sciences}, vol.~9, no.~18, p. 3897, 2019.

\bibitem{kim2019vision}
K.~Kim, A.~Jalal, and M.~Mahmood, ``Vision-based human activity recognition
  system using depth silhouettes: A smart home system for monitoring the
  residents,'' \emph{Journal of Electrical Engineering \& Technology}, vol.~14,
  no.~6, pp. 2567--2573, 2019.

\bibitem{buzzelli2020vision}
M.~Buzzelli, A.~Alb{\'e}, and G.~Ciocca, ``A vision-based system for monitoring
  elderly people at home,'' \emph{Applied Sciences}, vol.~10, no.~1, p. 374,
  2020.

\bibitem{khraief2020elderly}
C.~Khraief, F.~Benzarti, and H.~Amiri, ``Elderly fall detection based on
  multi-stream deep convolutional networks,'' \emph{Multimedia Tools and
  Applications}, pp. 1--24, 2020.

\bibitem{li2018learning}
H.~Li, K.~Ota, and M.~Dong, ``Learning iot in edge: Deep learning for the
  internet of things with edge computing,'' \emph{IEEE network}, vol.~32,
  no.~1, pp. 96--101, 2018.

\bibitem{chen2019deep}
J.~Chen and X.~Ran, ``Deep learning with edge computing: A review,''
  \emph{Proceedings of the IEEE}, vol. 107, no.~8, pp. 1655--1674, 2019.

\bibitem{wang2020convergence}
X.~Wang, Y.~Han, V.~C. Leung, D.~Niyato, X.~Yan, and X.~Chen, ``Convergence of
  edge computing and deep learning: A comprehensive survey,'' \emph{IEEE
  Communications Surveys \& Tutorials}, 2020.

\bibitem{wang2020generalizing}
Y.~Wang, Q.~Yao, J.~T. Kwok, and L.~M. Ni, ``Generalizing from a few examples:
  A survey on few-shot learning,'' \emph{ACM Computing Surveys (CSUR)},
  vol.~53, no.~3, pp. 1--34, 2020.

\bibitem{lecun2015deep}
Y.~LeCun, Y.~Bengio, and G.~Hinton, ``Deep learning,'' \emph{nature}, vol. 521,
  no. 7553, pp. 436--444, 2015.

\bibitem{esteva2019guide}
A.~Esteva, A.~Robicquet, B.~Ramsundar, V.~Kuleshov, M.~DePristo, K.~Chou,
  C.~Cui, G.~Corrado, S.~Thrun, and J.~Dean, ``A guide to deep learning in
  healthcare,'' \emph{Nature medicine}, vol.~25, no.~1, pp. 24--29, 2019.

\bibitem{tan2018survey}
C.~Tan, F.~Sun, T.~Kong, W.~Zhang, C.~Yang, and C.~Liu, ``A survey on deep
  transfer learning,'' in \emph{International conference on artificial neural
  networks}.\hskip 1em plus 0.5em minus 0.4em\relax Springer, 2018, pp.
  270--279.

\bibitem{sufian2020survey}
A.~Sufian, A.~Ghosh, A.~S. Sadiq, and F.~Smarandache, ``A survey on deep
  transfer learning to edge computing for mitigating the covid-19 pandemic,''
  \emph{Journal of Systems Architecture}, vol. 108, p. 101830, 2020.

\end{thebibliography}

\end{document}